\documentclass[10pt,twocolumn,letterpaper]{article}

\usepackage[pagenumbers]{cvpr}
\usepackage{fontspec}
\usepackage{booktabs}
\usepackage{array}
\usepackage{xcolor}
\usepackage{colortbl}
\usepackage{multirow}
\usepackage{graphicx}

\usepackage{tabularx}

\definecolor{cvprblue}{rgb}{0.21,0.49,0.74}

\title{
EviBench: Multi-Modal Product Understanding\\
under Missing and Conflicting Evidence
}

\author{\textbf{Meiqing Shan, Aiqing Zhuo, Yujia Wu}\\ \textbf{Libin Ma, Baokun Pan, Xing Gao}\\ Kuaishou Technology}

\usepackage{float}%

\definecolor{BestBlue}{RGB}{232,240,249}
\definecolor{BestText}{RGB}{57,88,119}
\definecolor{MethodBlue}{RGB}{246,249,253}

\definecolor{AblModelYellow}{RGB}{255,246,214}
\definecolor{AblBestBlue}{RGB}{225,239,250}

\usepackage{listings}
\usepackage{microtype}
\providecommand{\field}[1]{\texttt{#1}}
\usepackage[
    pagebackref,
    breaklinks,
    colorlinks,
    allcolors=cvprblue
]{hyperref}

\begin{document}

\maketitle

\raggedbottom

\begin{abstract}
E-commerce livestreams present product information through images, video, speech, and text, but these sources may be incomplete or inconsistent. Reliable multimodal product question answering requires assessing whether the available evidence supports an answer: a missing source does not necessarily make a question unanswerable, whereas insufficient evidence or an unresolved, relevant conflict calls for abstention. We introduce EviBench, a benchmark comprising 3,182 examples across five e-commerce question-answering tasks that jointly evaluates answering ability, evidence understanding, and abstention decisions. Evaluations of four open-source and six proprietary models show that selecting the correct answer does not imply accurately understanding the underlying evidence conditions: the model achieving the highest option accuracy of 86.83\% correctly identifies the evidence state on only 38.50\% of examples. To address this gap, we propose Evidence Intervention Group Relative Policy Optimization (EI-GRPO). Starting from supervised fine-tuning, EI-GRPO trains on paired inputs that share a question but differ in evidence conditions. It combines individual-response rewards, cross-condition field-correctness rewards, and separately computed advantages for each endpoint to coordinate answer selection with evidence assessment. Across four backbone models, EI-GRPO yields broader improvements in answer selection, evidence understanding, and appropriate abstention over supervised fine-tuning, reducing structured hallucinations without increasing over-refusal. These results demonstrate that EviBench reveals deficiencies in evidence assessment beyond answer correctness, while EI-GRPO helps models make more reliable answering decisions under missing and conflicting multimodal inputs.
\end{abstract}

\vspace{-6pt}
\section{Introduction}
\begin{figure}[t]
    \centering

    \includegraphics[width=1.0\linewidth]
    {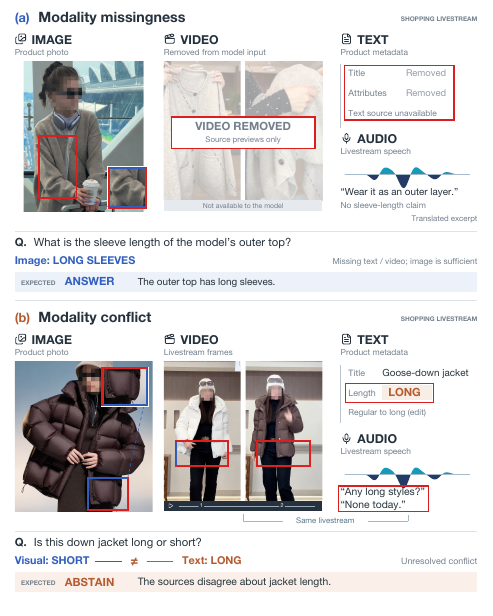}
    \caption{
        Missing and conflicting evidence in e-commerce livestreaming.
        (a) Visual evidence supports an answer despite missing sources.
        (b) Unresolved evidence conflict requires abstention.
        Red boxes highlight missing sources and decision-relevant evidence.
    }
    \label{fig:overview}
\end{figure}

E-commerce livestreaming provides complementary product views through images, video, and text, as studied in livestream product recognition~\cite{yang2023lpr4m}. Together with speech and metadata, these sources may be incomplete~\cite{lee2023missing} or inconsistent~\cite{zhang2025modalityconflict}. Multimodal large language models (MLLMs) must determine whether this evidence supports an answer~\cite{whitehead2022reliablevqa}.
In Fig.~\ref{fig:overview}(a), the image supports long sleeves
despite missing video and metadata. In Fig.~\ref{fig:overview}(b),
the image and metadata disagree on jacket length, requiring
abstention if the relevant conflict remains unresolved.
Poor-quality images can also make real-world questions unanswerable~\cite{gurari2018vizwiz}. Reliable answering requires assessing both evidence sufficiency
and consistency.

Existing benchmarks study e-commerce visual utility~\cite{ling2025ecommmmu},
cross-modal contradictions~\cite{popordanoska2025clash}, and
abstention under insufficient or corrupted evidence~\cite{madhusudhan2026mmaqa,nazi2026omdbench}.
VQA-CP~\cite{agrawal2018vqacp}, HallusionBench~\cite{guan2024hallusionbench}, and POPE~\cite{li2023pope} expose reliance on language or object co-occurrence priors rather than visual evidence. We jointly evaluate these behaviors in product QA, where missingness
and conflict can coexist. Answer accuracy alone cannot reveal
unsupported citations, missed contradictions, or unnecessary refusal.

We introduce EviBench, a benchmark of 3,182 human-verified
examples spanning five e-commerce QA tasks.
Controlled source removal, visual evidence reduction, and
contradictory textual edits produce conditions ranging from
complete and consistent evidence to combined missingness
and conflict.
Our structured protocol evaluates answer selection, evidence
states, source attribution, conflict detection, and abstention,
alongside structured hallucination and over-refusal.
Evaluation of four open-source and six commercial models
reveals gaps between answer correctness and evidence assessment:
the model with the highest option accuracy achieves 86.83\%
but only 38.50\% evidence-state accuracy.

We propose Evidence Intervention Group Relative Policy Optimization
(EI-GRPO), building on visual-intervention grounding methods~\cite{xu2025defacto,yu2026cfpo}.
After SFT, EI-GRPO trains on two evidence versions of the same
question with endpoint-specific references and advantages,
combining single-response rewards with cross-condition
field-correctness bonuses. Models learn to preserve supported
answers, revise them when decisive evidence changes, and abstain
when required. Across four backbones, EI-GRPO improves answer
selection, source attribution, and conflict detection while
reducing structured hallucination without increasing aggregate
over-refusal relative to SFT.

Our contributions are threefold:
\begin{itemize}
    \item \textbf{Benchmark and evaluation.}
    EviBench jointly evaluates product QA, evidence attribution,
    and abstention under controlled missing and conflicting
    evidence.

    \item \textbf{Evidence intervention training.}
    EI-GRPO combines cross-condition field rewards with
    endpoint-specific credit assignment to improve decisions
    on paired evidence inputs.

    \item \textbf{Empirical analysis.}
    Evaluations of ten MLLMs and post-training experiments on
    four backbones reveal evidence-reliability gaps and
    demonstrate the benefits of EI-GRPO.
\end{itemize}

\section{Related Work} \subsection{Multimodal Large Language Models} LLaVA~\cite{liu2023llava} develops visual instruction tuning, while Qwen2.5-Omni~\cite{qwen2025omni} extends unified processing to text, images, audio, and video. Broader input coverage, however, does not ensure evidence-grounded generation. M3ID~\cite{favero2024m3id} links hallucinations to declining visual dependence during generation and amplifies image-conditioned predictions relative to language priors. PAPO~\cite{wang2026papo} instead compares predictions under original and masked images during policy optimization. mDPO~\cite{wang2024mdpo} identifies a related failure: response preferences can be learned while ignoring the image. Its image-preference objective and reward anchor constrain both visual dependence and chosen-response likelihood, showing why response-level preference alone can leave grounding underdetermined. Feedback granularity also matters: LLaVA-RLHF~\cite{sun2024factrlhf} uses factual annotations in reward modeling, whereas RLHF-V~\cite{yu2024rlhfv} learns from segment-level corrections. These decoding- and training-time approaches make visual dependence explicit, treating grounding as a property of how predictions respond to evidence rather than answer correctness alone. \subsection{Modality Conflict and Missingness} Missing-modality methods adapt to available inputs or reconstruct absent content: Lee et al.~\cite{lee2023missing} use missing-aware prompts for frozen transformers, whereas Knowledge Bridger~\cite{ke2025bridger} generates and ranks missing modalities using structured knowledge. Reconstruction thus introduces an additional requirement: inferred content must remain consistent with observed evidence. Under conflict, Words or Vision~\cite{deng2025wordsvision} identifies text preference over contradictory images, and MMMC~\cite{zhang2025modalityconflict} compares prompting, supervised tuning, and reinforcement learning. MAD~\cite{chung2026mad} addresses audio--visual interference by weighting modality-specific decoding branches using self-assessed task relevance. Together with attention steering in EmoMM~\cite{sun2026emomm}, it shifts the focus from merely adding modalities to controlling their influence. MoD-DPO~\cite{chaubey2026moddpo} makes this control explicit during preference learning, enforcing invariance to irrelevant-modality corruption and sensitivity to relevant-modality perturbations, with an additional language-prior penalty. Its distinction between relevant and irrelevant changes makes robustness conditional on the question's evidence requirements. Such control must also permit justified answer changes: BPO~\cite{kang2026bpo} jointly trains stability, sensitivity, and abstention across counterfactual video views, rather than enforcing unconditional consistency. \subsection{Multimodal Benchmarks} EcomMMMU~\cite{ling2025ecommmmu} shows that additional product images can introduce noise and redundancy. AVHBench~\cite{kim2025avhbench} separately tests audio-induced visual hallucinations and visual-induced audio hallucinations. This directional evaluation reveals whether a model confuses evidence that an event is audible with evidence that it is visible, making the origin of cross-modal errors observable. CLASH~\cite{popordanoska2025clash}, MM-AQA~\cite{madhusudhan2026mmaqa}, and OMD-Bench~\cite{nazi2026omdbench} probe complementary failures through contradictions, insufficient evidence, and controlled modality corruption. Reliable VQA~\cite{whitehead2022reliablevqa} measures the risk--coverage trade-off, while Cha et al.~\cite{cha2026visualanswerability} test answer-preserving, evidence-removing, and answer-changing interventions. Javadov et al.~\cite{javadov2026missingness} further compare models' stated evidence sufficiency with their behavior after restoring missing modalities, finding systematic overestimation of sufficiency. This separates a model's claimed reliance from its observed reliance: plausible evidence explanations need not predict actual behavior. Collectively, these protocols support evaluating answer accuracy, abstention, and evidence dependence together.%

\begin{figure}[!htbp]
    \centering
    \includegraphics[width=\linewidth]
    {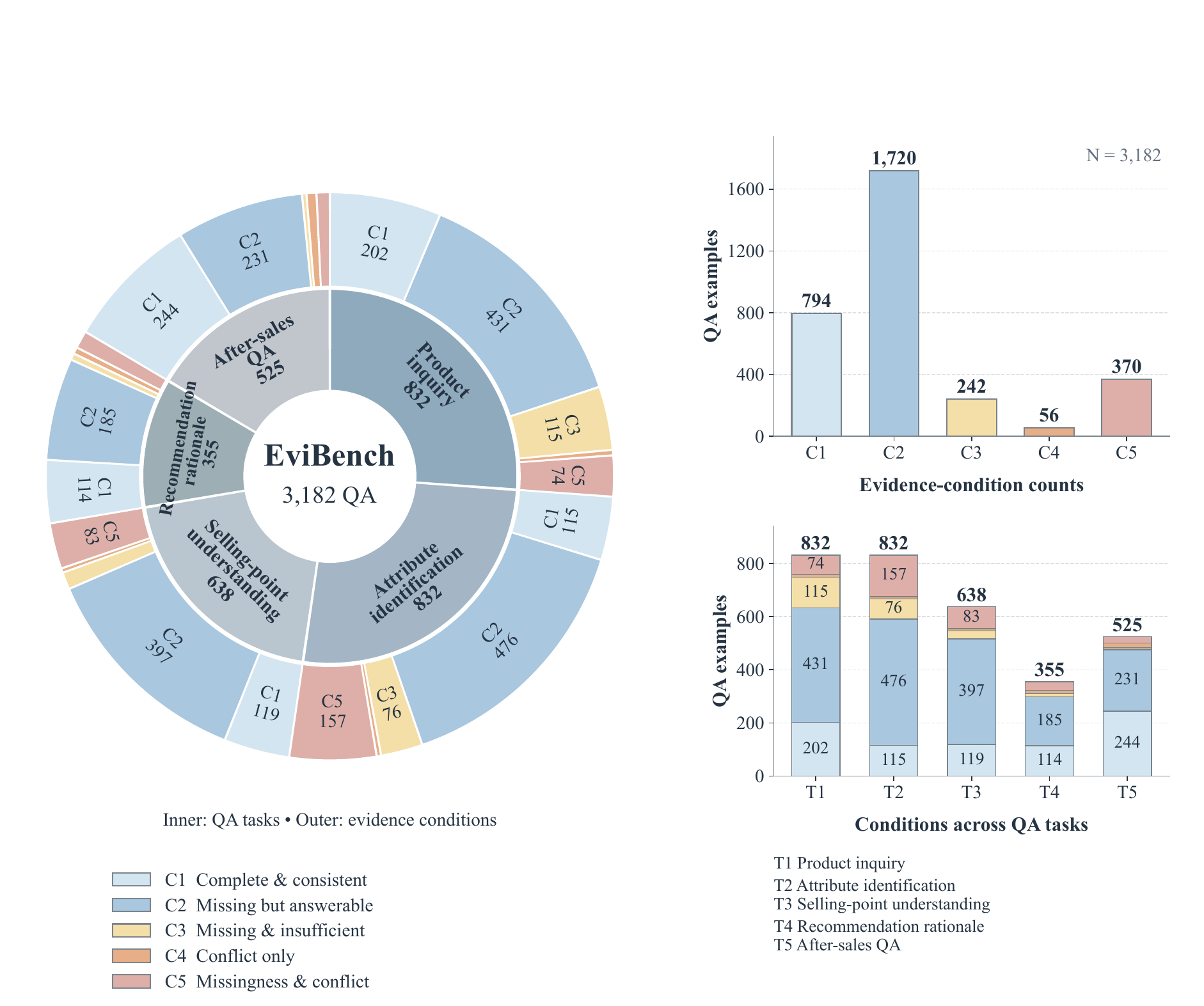}
    \caption{
        Sample counts across five QA tasks.
    }
    \label{fig:task_distribution}
\end{figure}

\section{The EviBench Benchmark}
\label{sec:evibench}
\label{sec:problem_formulation}

EviBench contains 3,182 questions evaluating evidence-grounded
multimodal decisions under missingness and conflict. It covers
product inquiry, attribute identification, selling-point
understanding, recommendation rationale, and after-sales QA
(Fig.~\ref{fig:task_distribution}).

\subsection{Data Construction}
\label{sec:data_construction}

\noindent\textbf{Data generation and verification.}
Model-generated instructions can expand supervision~\cite{wang2023selfinstruct}. Starting from 1,160 full-modality examples containing product
images, livestream video and audio, ASR transcripts, titles,
and attributes, we use GPT-5.5 and Gemini 3.1 Pro to generate
questions, four candidate responses, and reference decisions.
Synthetic instruction data can contain hallucinations~\cite{yu2024hallucidoctor}.
We verify every generated example: 93.7\%
are accepted without modification, and the remainder are
corrected before inclusion. The supplementary material details the human
verification protocol, including the criteria for
evidence sufficiency and unresolved conflicts,
reviewer responsibilities, and quality-control
procedures.

\noindent\textbf{Evidence interventions.}
We remove sources or reduce visual evidence to construct missingness, retaining answerable labels only when sufficient support remains. Conflicts are introduced by editing product text to contradict visual, spoken, or other textual evidence about the queried fact. Only unresolved relevant contradictions require conflict-based abstention. Combining source removal and contradictory edits produces missing-and-conflicting examples, denoted M+C. The benchmark includes 2,332 examples with missingness and 426 with conflict, overlapping in 370 examples. Its five conditions are Complete, with 794 complete-and-consistent examples; M-A, with 1,720 missing-but-answerable examples; M-I, with 242 missing-and-insufficient examples; Conflict, with 56 conflict-only examples; and M+C, with 370 missing-and-conflicting examples. Separately, we construct 7,996 SFT examples and a reinforcement-learning pool of 7,897 original questions. Data provenance, sample construction counts,
split procedures, overlap checks, and counts of
unique products, livestreams, and intervention groups
are provided in the supplementary material.

\noindent\textbf{Input and output.}
Each input $x=(q,O,E)$ comprises a question, four candidate
responses, and available evidence. The prompt specifies source
availability and JSON-formatted text, accompanied by up to four
images, eight video frames, and the full supplied audio for
audio-capable models. Reference annotations are excluded.
Models return a structured JSON decision:

\begin{equation}
    \hat y=(\hat o,\hat z,\hat v,\hat E,\hat C,\hat a,\hat c),
    \label{eq:structured_decision}
\end{equation}
where the fields denote the selected option, evidence state,
factual value, cited sources, conflict source pair, abstention,
and confidence, respectively. States are \texttt{supported}
(complete and sufficient), \texttt{missing} (missing but sufficient),
\texttt{insufficient}, or \texttt{conflict}. The latter two
require abstention and an empty factual value; unresolved
conflict takes precedence over missingness.
Both conflict-only and missing-plus-conflict conditions
map to the \texttt{conflict} state when the contradiction
remains unresolved.

\subsection{Evaluation Metrics}
\label{sec:evaluation_protocol}

Rationale evaluation distinguishes annotation agreement from faithfulness~\cite{deyoung2020eraser}. We evaluate answer selection, evidence assessment, source
attribution, and abstention separately. Let $N=3{,}182$,
let $y_i^{*}$ denote the reference, and let $\mathbf{1}[\cdot]$
be the indicator function. All scores are defined as
proportions and reported as percentages.

\noindent\textbf{Option and state accuracy.}
Option accuracy measures correct response selection, whereas
state accuracy measures recognition of the four evidence states:

\begin{equation}
    \mathrm{OptionAcc}
    =\frac{1}{N}\sum_{i=1}^{N}
    \mathbf{1}[\hat o_i=o_i^{*}].
    \label{eq:option_accuracy}
\end{equation}

\begin{equation}
    \mathrm{StateAcc}
    =\frac{1}{N}\sum_{i=1}^{N}
    \mathbf{1}[\hat z_i=z_i^{*}].
    \label{eq:state_accuracy}
\end{equation}

\begin{figure}[!htbp]
    \centering
    \includegraphics[width=\linewidth]
    {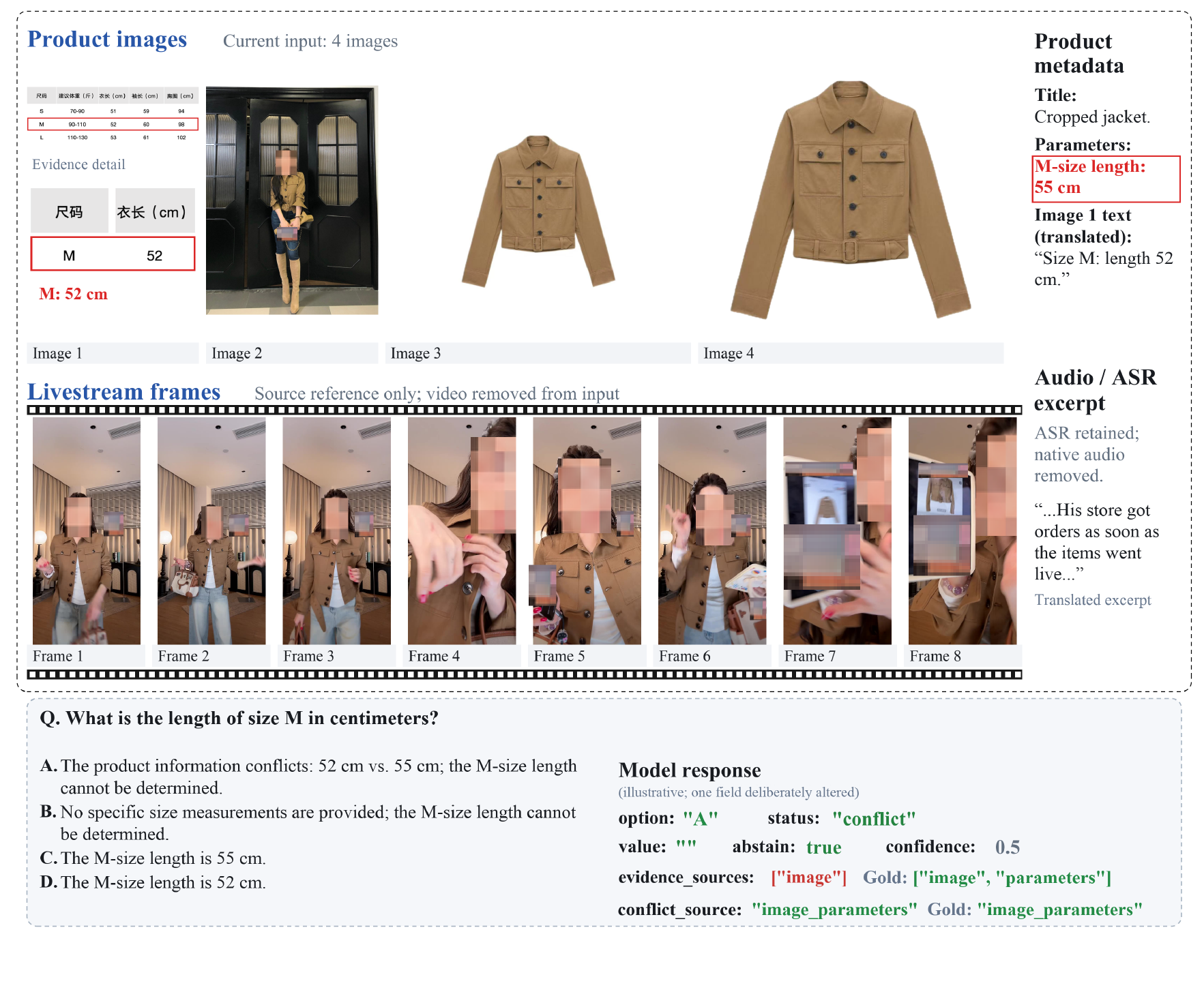}
    \caption{
        An illustrative image--parameter conflict:
        the size chart reports 52 cm, while edited parameters
        specify 55 cm. The response correctly detects the
        conflict and abstains but omits the parameter citation.
        Video frames are reference-only; faces and brand
        identifiers are mosaicked.
    }
    \label{fig:sample_example}
\end{figure}

\noindent\textbf{Source F1.}
Citation benchmarks assess answer correctness and citation quality separately~\cite{gao2023alce}. This metric measures agreement between predicted citations
and annotated evidence sources. Let $\mathcal G_i$ contain
the accepted reference source sets, including annotated
alternatives. We take the best set match for each example
and then average across examples:
\begin{equation}
    \mathrm{SourceF1}
    =\frac{1}{N}\sum_{i=1}^{N}
    \max_{G\in\mathcal G_i}
    \frac{2|\hat E_i\cap G|}{|\hat E_i|+|G|}.
    \label{eq:source_f1}
\end{equation}
Two empty sets receive a score of one; an invalid source-list
format receives zero. Additional unsupported citations reduce
precision. On conflict examples, reference citations include
the sources establishing the contradiction.
In Fig.~\ref{fig:sample_example}, citing only \texttt{image}
rather than \{\texttt{image},\texttt{parameters}\} gives Source F1
$2/3$ ($66.67\%$); a correct \texttt{conflict\_source} does not
compensate for the omitted citation.

\noindent\textbf{Conflict F1.}
This metric evaluates binary conflict detection across all
examples. A prediction claims conflict if its state is
\texttt{conflict} or it specifies a valid nonempty conflict
source pair. With $\mathrm{TP}$, $\mathrm{FP}$, and
$\mathrm{FN}$ defined against the reference conflict labels,
\begin{equation}
    \mathrm{ConflictF1}
    =\frac{2\mathrm{TP}}
    {2\mathrm{TP}+\mathrm{FP}+\mathrm{FN}}.
    \label{eq:conflict_f1}
\end{equation}
Conflict F1 uses dataset-level counts and does not measure
source-pair localization. The conflict prediction in
Fig.~\ref{fig:sample_example} contributes one true positive.

\noindent\textbf{Abstention and over-refusal.}
Let $\mathcal R=\{i:a_i^{*}=1\}$ and
$\mathcal A=\{i:a_i^{*}=0\}$, with
$|\mathcal R|=668$ and $|\mathcal A|=2{,}514$.
Correct Abstention Rate (CAR) measures abstention when required;
Over-refusal Rate (OR) measures abstention despite sufficient
evidence:

\begin{equation}
    \mathrm{CAR}
    =\frac{1}{|\mathcal R|}
    \sum_{i\in\mathcal R}\mathbf{1}[\hat a_i=1].
    \label{eq:correct_abstention_rate}
\end{equation}

\begin{equation}
    \mathrm{OR}
    =\frac{1}{|\mathcal A|}
    \sum_{i\in\mathcal A}\mathbf{1}[\hat a_i=1].
    \label{eq:over_refusal_rate}
\end{equation}

Higher CAR and lower OR indicate better answerability
decisions. CAR evaluates the abstention field, not the
correctness of every output field.

\noindent\textbf{Structured hallucination rate.}
CHAIR~\cite{rohrbach2018chair} measures object hallucination, while AMBER~\cite{wang2023amber} also tests attributes and relations. Here, we measure three types of evidence-related violations
that can be determined from structured predictions
and reference annotations.
Let $\mathcal V$ denote the set of schema-valid outputs.
For each $i\in\mathcal V$, $u_i$ indicates that the
reference requires abstention but the prediction sets
\texttt{abstain=false} or provides a nonempty
\texttt{value}; $s_i$ indicates that
\texttt{evidence\_sources} includes a source marked
unavailable in the evaluated input; and $w_i$ indicates
a nonempty incorrect \texttt{value} on a sample with
a scored reference value.
We count each output at most once:
\begin{equation}
    \mathrm{HR}
    =\frac{1}{|\mathcal V|}
    \sum_{i\in\mathcal V}
    \mathbf{1}[u_i \lor s_i \lor w_i].
    \label{eq:structured_hr}
\end{equation}
This metric captures structured violations rather than
an exhaustive semantic assessment of the selected
candidate response or any additional text.
In particular, an incorrect option is not automatically
classified as a hallucination, and citing an available
but irrelevant source is not fully covered.
Schema-invalid outputs are excluded from the denominator,
rather than treated as hallucination-free.
We therefore report the coverage
$\mathrm{Schema}=|\mathcal V|/N$ alongside HR and
interpret HR together with over-refusal rate to account for the reliability--coverage trade-off~\cite{whitehead2022reliablevqa}.

In Fig.~\ref{fig:sample_example}, abstaining with an empty value
and only available citations gives $u_i=s_i=w_i=0$, despite
incomplete attribution. Returning either measurement would set $u_i=1$.

\begin{figure*}[t]
    \centering

    \includegraphics[width=\textwidth]
    {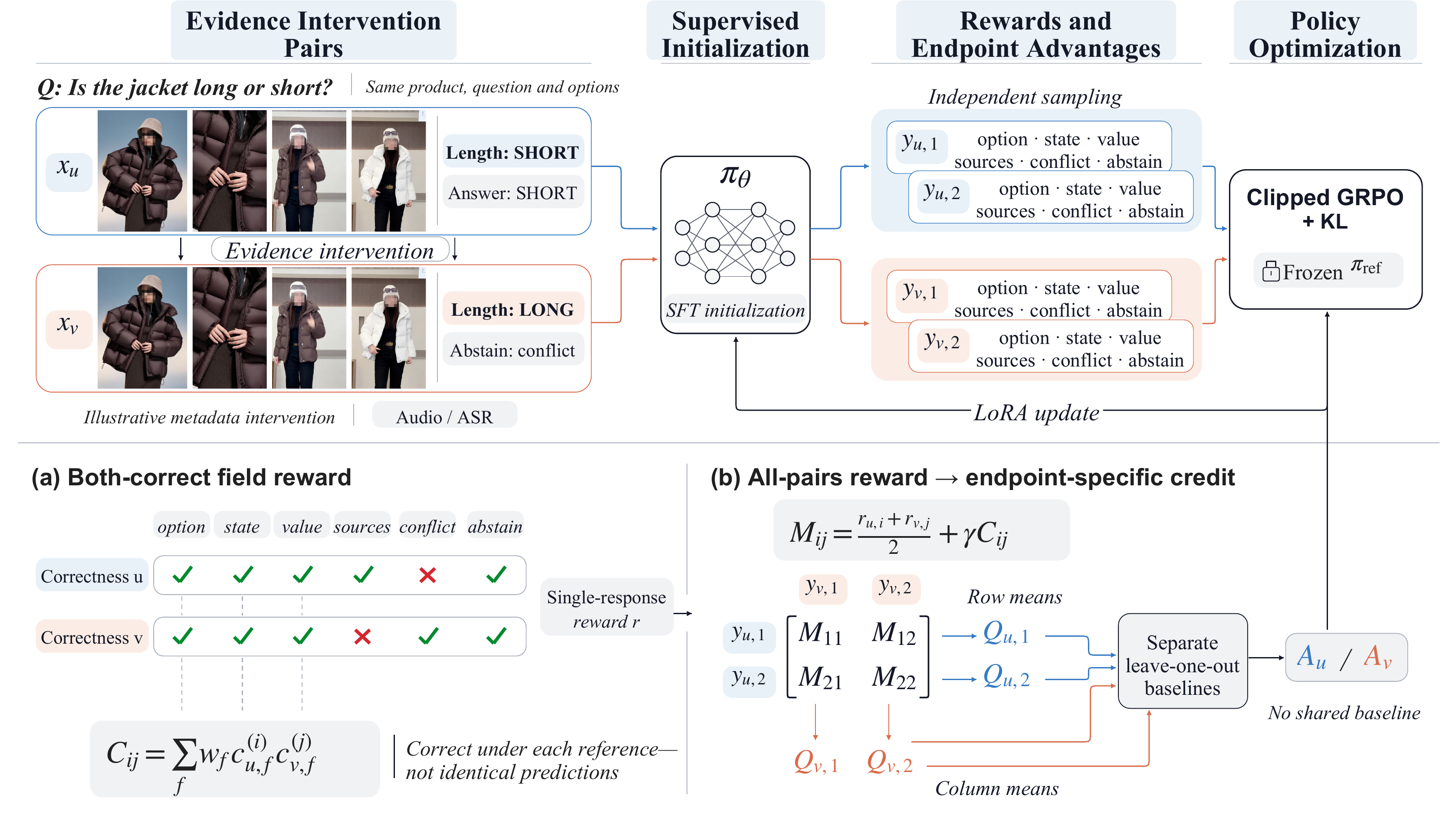}

    \caption{Overview of Evidence Intervention GRPO (EI-GRPO).}
    \label{fig:ei_grpo_method}
\end{figure*}

\section{Evidence Intervention Group Relative Policy Optimization}

EI-GRPO extends GRPO~\cite{shao2024deepseekmath}
to two evidence versions of the same product question.
Following supervised initialization, it combines single-response
rewards, cross-endpoint field bonuses, and endpoint-specific
advantages (Fig.~\ref{fig:ei_grpo_method}) to learn when to
preserve, revise, or abstain from an answer.

\subsection{Supervised Initialization}
  \label{sec:training_initialization}

  Following prior RLHF practice~\cite{ouyang2022instructgpt}, we use SFT to learn the
  structured response format in Sec.~3.1.
  For input $x$ and reference response $y^{*}$,
  $\mathcal{T}$ contains supervised response positions,
  excluding the prompt, media placeholders, and numerical
  confidence placeholder.
  We minimize the average negative log-likelihood~\cite{liu2023llava}:
  \begin{equation}
  \label{eq:sft}
  \mathcal{L}_{\mathrm{SFT}}
  =
  -\mathbb{E}_{(x,y^{*})}
  \left[
  \frac{1}{|\mathcal{T}|}
  \sum_{t \in \mathcal{T}}
  \log \pi_{\theta}
  \left(y_t^{*} \mid x, y_{<t}^{*}\right)
  \right].
  \end{equation}
  The SFT checkpoint initializes both $\pi_{\theta}$ and
  $\pi_{\mathrm{ref}}$.
  EI-GRPO updates $\pi_{\theta}$ on paired evidence inputs
  while keeping $\pi_{\mathrm{ref}}$ frozen for regularization.

  \subsection{Evidence Intervention Pairs}
  \label{sec:intervention_pairs}

  An intervention pair $(x_u,x_v)$ shares the same product,
  question, and ordered response options, differing only
  in evidence.
  Removing redundant support or introducing irrelevant
  discrepancies preserves the supported answer.
  Removing the last sufficient support or introducing an
  unresolved relevant conflict requires abstention;
  replacing decisive evidence may change the answer~\cite{cha2026visualanswerability}.

  Each endpoint has reference labels for six fields:
  option, evidence state, factual value, cited sources,
  conflicting sources, and abstention.
  Let $c_{s,f}(y)\in\{0,1\}$ denote correctness of field
  $f\in\mathcal{F}$ at endpoint $s\in\{u,v\}$, with
  $c_{s,f}^{(i)}=c_{s,f}(y_{s,i})$ for candidate $i$.
  Cited sources must exactly match an accepted reference
  set; unsupported additions invalidate the match.
  Malformed responses receive no field credit.
  These indicators determine rewards, not additional
  supervised losses.

\subsection{Rewards and Endpoint Advantages}
\label{sec:cross_endpoint_rewards}

We first evaluate each response against its own endpoint reference. The base reward combines option correctness, joint correctness of the evidence state and abstention decision under the consistency rules, source-set F1, conflict-source pair correctness, factual value correctness, and complete-response correctness:

\begin{equation}
\label{eq:base_reward}
r_{\mathrm{base}}(x,y)
=
\sum_{k\in\mathcal{K}}
\lambda_k r_k(x,y),
\qquad \lambda_k\geq 0,
\end{equation}

where $\mathcal{K}=
\{\mathrm{opt},\mathrm{sta},\mathrm{src},
\mathrm{con},\mathrm{val},\mathrm{all}\}$.
The source component uses the best F1 over accepted source sets, whereas the source-field indicator in Sec.~\ref{sec:intervention_pairs} requires an exact match.

We adjust this reward to penalize invalid outputs and unsupported decisions.
A malformed response receives $-\eta_{\mathrm{fmt}}$.
For a valid response, citing any unavailable source incurs a penalty of $\eta_{\mathrm{src}}$.
We then cap the reward at $\tau_{\mathrm{over}}$ for abstaining on an answerable question, or at $\tau_{\mathrm{unsafe}}$ for answering or returning
a nonempty factual value when abstention is required.
Here, $\eta_{\mathrm{fmt}},\eta_{\mathrm{src}}>0$ and $\tau_{\mathrm{unsafe}}<\tau_{\mathrm{over}}\leq 0$.
Numerical confidence does not contribute to the reward. We denote the adjusted reward of candidate $y_{s,i}$ by $r_{s,i}$.

We independently sample $K_u$ and $K_v$ responses at the two endpoints and score all $K_uK_v$ combinations.
Each combination receives the mean single-response reward plus a bonus for fields that are correct under both evidence conditions:

\begin{align}
\label{eq:joint_reward}
M_{ij}
&=
\frac{r_{u,i}+r_{v,j}}{2}
+
\gamma C_{ij}.
\\
\label{eq:joint_field_bonus}
C_{ij}
&=
\sum_{f\in\mathcal{F}}
w_f\,c_{u,f}^{(i)}c_{v,f}^{(j)}.
\end{align}

Here, $\gamma$ controls the joint bonus and $w_f$ weights each field.
A field earns the bonus only when both responses satisfy their respective references; their predicted values need not be identical.
Thus, a supported answer at one endpoint and a required abstention at the other can jointly receive credit.

Pairing candidates only by sampling index makes credit depend on an arbitrarily selected partner.
Instead, we average each candidate's rewards over all opposite-endpoint candidates and apply a leave-one-out baseline~\cite{ahmadian2024reinforce} within its own endpoint.
For endpoint $u$, we compute

\begin{align}
\label{eq:partner_mean_reward}
Q_{u,i}
&=
\frac{1}{K_v}
\sum_{j=1}^{K_v}M_{ij}.
\\
\label{eq:endpoint_advantage}
A_{u,i}
&=
Q_{u,i}
-
\frac{1}{K_u-1}
\sum_{\substack{\ell=1\\\ell\neq i}}^{K_u}
Q_{u,\ell}.
\end{align}

Endpoint $v$ uses columns symmetrically, with $K_u,K_v\geq 2$.
The partner's mean single-response reward cancels under baseline subtraction, while its field correctness still affects the joint bonus.
Single-response rewards can distinguish candidates when the joint bonus cannot.
We do not normalize advantages by the group standard deviation.

\subsection{Policy Optimization}
\label{sec:endpoint_policy_optimization}

We optimize the policy with the PPO clipped surrogate~\cite{schulman2017ppo} and a sampled-token reference penalty as in GRPO~\cite{shao2024deepseekmath}.
For each token, let $p_{\theta}$, $p_{\mathrm{old}}$, and $p_{\mathrm{ref}}$ denote its probabilities conditioned on the input and generated prefix.
Define $\rho=p_{\theta}/p_{\mathrm{old}}$, $\bar{\rho}=\operatorname{clip}(\rho,1-\epsilon,1+\epsilon)$, and $d=\exp(\delta)-\delta-1$, where $\delta=\log p_{\mathrm{ref}}-\log p_{\theta}$.
The old policy determines the importance ratio, while the frozen SFT reference provides regularization.

For each intervention pair, the objective averages over optimized tokens within each response and then equally over all $K=K_u+K_v$ responses:

\begin{gather}
\label{eq:ei_objective}
\mathcal{L}_{\mathrm{EI}}
=
\frac{1}{K}
\sum_{s\in\{u,v\}}
\sum_{i=1}^{K_s}
\frac{1}{n_{s,i}}
\sum_{t=1}^{n_{s,i}}
\ell_{s,i,t}.
\\
\label{eq:token_loss}
\begin{aligned}
\ell_{s,i,t}
={}&-\min\!\left(
\rho_{s,i,t}A_{s,i},
\bar{\rho}_{s,i,t}A_{s,i}
\right) \\
&+\beta d_{s,i,t}.
\end{aligned}
\end{gather}

Here, $n_{s,i}$ is the number of optimized tokens in response $y_{s,i}$, and each token uses its response's endpoint-specific advantage.
The clipping parameter $\epsilon$, reference coefficient $\beta$, and reward hyperparameters are specified in the experimental setup.
This objective requires neither a critic nor a learned reward model.

\subsection{Experimental Setup}
\label{sec:experimental_setup}

\noindent\textbf{Models.}
We evaluate ten representative MLLMs on EviBench, including four open-source
models: Qwen2.5-Omni-7B~\cite{qwen2025omni}, Qwen3-Omni-30B~\cite{xu2025qwen3omni}, TLive-Omni-4B and TLive-Omni-9B~\cite{hu2026tliveomni}, and six commercial models:
Gemini 3.1 Pro, Gemini 3 Flash, GPT-5.2, GPT-5.4,
GPT-5.5, and Claude Opus 4.7.
For post-training experiments, we compare EI-GRPO with
DPO~\cite{rafailov2023dpo},
standard GRPO~\cite{shao2024deepseekmath},
mDPO~\cite{wang2024mdpo},
and BPO~\cite{kang2026bpo}.

\textbf{Implementation Details.}
Training uses Transformers~\cite{wolf2020transformers} and SWIFT~\cite{zhao2025swift}. SFT uses 7,996 examples and LoRA~\cite{hu2022lora} on language-attention layers with frozen multimodal encoders, learning rate $10^{-5}$, rank 16, scaling 32, and dropout 0.05. All five post-training methods share 7,897 original questions. Samples linked by product, livestream, or intervention are grouped within one split; the 3,408-example validation pool and EviBench test set come from distinct source partitions.
Base-reward weights in Eq.~\ref{eq:base_reward}, ordered as $(\mathrm{opt},\mathrm{sta},\mathrm{src},\mathrm{con},\mathrm{val},\mathrm{all})$, are $(0.20,0.25,0.20,0.10,0.10,0.15)$. Validation sweeps of response count, source-bonus weight, and reference regularization yield $K_u=K_v=2$, $w_{\mathrm{src}}=0.40$, and $\beta=0.10$, balancing performance and cost. We set $\gamma=0.25$ and $\epsilon=0.20$; further details are in the supplementary material.

\begin{table*}[!t]
\centering
\caption{Prompt-only benchmark results without task-specific
training. All scores are percentages. Best results are bolded
separately within the closed-source and open-source groups.}
\label{tab:base_response_evidence}

\footnotesize
\setlength{\tabcolsep}{4pt}
\renewcommand{\arraystretch}{1.0}

\begin{tabular*}{\textwidth}{
    @{\extracolsep{\fill}}lccccccc@{}
}
\toprule
\textbf{Model}
& \textbf{Opt.}$\uparrow$
& \textbf{State}$\uparrow$
& \textbf{Source F1}$\uparrow$
& \textbf{Conf. F1}$\uparrow$
& \textbf{HR}$\downarrow$
& \textbf{CAR}$\uparrow$
& \textbf{OR}$\downarrow$ \\
\midrule

\rowcolor{MethodBlue}
\multicolumn{8}{l}{\textbf{Closed-source models}} \\

Gemini 3.1 Pro
& 84.10
& \textbf{76.59}
& \textbf{80.54}
& 94.53
& 18.14
& 96.56
& 20.21 \\

Gemini 3 Flash
& 78.28
& 59.40
& 76.92
& 95.05
& \textbf{11.27}
& 93.26
& 22.28 \\

GPT-5.2
& 78.28
& 31.49
& 78.16
& 77.47
& 22.12
& 74.55
& 21.44 \\

GPT-5.4
& 82.08
& 62.38
& 76.68
& 96.63
& 15.57
& 96.11
& 20.80 \\

GPT-5.5
& \textbf{86.83}
& 38.50
& 76.55
& \textbf{96.93}
& 15.30
& \textbf{97.01}
& \textbf{16.19} \\

Claude Opus 4.7
& 83.63
& 44.39
& 74.46
& 93.00
& 12.71
& 91.02
& 19.94 \\

\midrule

\rowcolor{MethodBlue}
\multicolumn{8}{l}{\textbf{Open-source models}} \\

Qwen2.5-Omni-7B
& 68.54
& 23.29
& 46.13
& 13.20
& 32.33
& 15.87
& 26.15 \\

Qwen3-Omni-30B
& \textbf{76.08}
& \textbf{26.43}
& \textbf{76.01}
& 14.72
& 28.01
& 22.60
& 27.20 \\

TLive-Omni-4B
& 73.70
& 24.54
& 64.16
& 36.64
& 23.33
& 47.31
& \textbf{22.95} \\

TLive-Omni-9B
& 65.52
& 23.26
& 60.37
& \textbf{51.22}
& \textbf{21.49}
& \textbf{55.69}
& 32.82 \\

\bottomrule
\end{tabular*}
\end{table*}

\begin{table*}[!t]
\centering
\caption{Post-training results (\%) on four backbones, each sharing an SFT checkpoint across methods. Best results (including ties) are bold. Opt./State: option/state accuracy; Src./Conf.: source/conflict F1.}
\label{tab:training_comparison}

\footnotesize
\setlength{\tabcolsep}{1.5pt}
\renewcommand{\arraystretch}{1.0}

\begin{minipage}[t]{0.49\textwidth}
\centering
\begin{tabular*}{\linewidth}{
    @{\extracolsep{\fill}}lrrrrrrr@{}
}
\toprule
\multicolumn{8}{c}{\textbf{Qwen2.5-Omni-7B}} \\
\midrule
Method
& Opt.$\uparrow$ & State$\uparrow$
& Src.$\uparrow$ & Conf.$\uparrow$
& HR$\downarrow$ & CAR$\uparrow$ & OR$\downarrow$ \\
\midrule
SFT
& 72.06 & 56.29 & 50.06 & 45.20 & 20.11 & 52.99 & 22.51 \\
DPO
& 70.58 & 54.02 & 57.53 & 34.10 & 22.41 & 45.66 & 22.99 \\
GRPO
& 72.00 & 55.66 & 51.68 & 38.20 & 21.28 & 48.65 & 21.80 \\
mDPO
& 69.33 & 53.80 & 57.03 & 43.56 & 20.64 & 53.14 & 26.37 \\
BPO
& 72.92 & 55.42 & 56.74 & 49.30 & 20.31 & 55.54 & 20.76 \\
\rowcolor{MethodBlue}
\textbf{EI-GRPO}
& \textbf{75.33} & \textbf{64.66}
& \textbf{69.93} & \textbf{58.87}
& \textbf{18.73} & \textbf{59.13} & \textbf{15.63} \\
\bottomrule
\end{tabular*}
\end{minipage}\hfill%
\begin{minipage}[t]{0.49\textwidth}
\centering
\begin{tabular*}{\linewidth}{
    @{\extracolsep{\fill}}lrrrrrrr@{}
}
\toprule
\multicolumn{8}{c}{\textbf{Qwen3-Omni-30B}} \\
\midrule
Method
& Opt.$\uparrow$ & State$\uparrow$
& Src.$\uparrow$ & Conf.$\uparrow$
& HR$\downarrow$ & CAR$\uparrow$ & OR$\downarrow$ \\
\midrule
SFT
& 84.48 & 70.77 & 61.91 & 84.89 & 14.06 & 83.23 & 13.40 \\
DPO
& 83.56 & 67.06 & 66.31 & 73.10 & 16.77 & 72.16 & 12.05 \\
GRPO
& 84.51 & 69.70 & 62.92 & 85.16 & 13.79 & 84.43 & 14.64 \\
mDPO
& 84.00 & 70.30 & 65.41 & 82.99 & 14.37 & 81.59 & 14.32 \\
BPO
& 84.60 & 70.24 & 63.25 & 85.19 & 14.16 & 84.28 & 14.20 \\
\rowcolor{MethodBlue}
\textbf{EI-GRPO}
& \textbf{87.14} & \textbf{78.67}
& \textbf{67.54} & \textbf{88.64}
& \textbf{12.39} & \textbf{89.13} & \textbf{11.20} \\
\bottomrule
\end{tabular*}
\end{minipage}

\par\medskip

\begin{minipage}[t]{0.49\textwidth}
\centering
\begin{tabular*}{\linewidth}{
    @{\extracolsep{\fill}}lrrrrrrr@{}
}
\toprule
\multicolumn{8}{c}{\textbf{TLive-Omni-4B}} \\
\midrule
Method
& Opt.$\uparrow$ & State$\uparrow$
& Src.$\uparrow$ & Conf.$\uparrow$
& HR$\downarrow$ & CAR$\uparrow$ & OR$\downarrow$ \\
\midrule
SFT
& 75.61 & 66.37 & 68.55 & 91.37 & 12.63 & 92.22 & 29.08 \\
DPO
& 71.65 & 66.56 & 68.72 & 74.12 & 16.67 & 74.10 & 28.92 \\
GRPO
& 74.83 & 66.75 & 68.38 & 88.94 & 13.12 & 88.92 & 29.32 \\
mDPO
& 72.03 & 67.54 & 68.31 & 78.98 & 15.48 & 78.59 & 29.67 \\
BPO
& 75.61 & 66.40 & 68.79 & 90.53 & 12.69 & 90.87 & 29.00 \\
\rowcolor{MethodBlue}
\textbf{EI-GRPO}
& \textbf{78.27} & \textbf{76.12}
& \textbf{71.37} & \textbf{93.04}
& \textbf{11.76} & \textbf{98.80} & \textbf{27.28} \\
\bottomrule
\end{tabular*}
\end{minipage}\hfill%
\begin{minipage}[t]{0.49\textwidth}
\centering
\begin{tabular*}{\linewidth}{
    @{\extracolsep{\fill}}lrrrrrrr@{}
}
\toprule
\multicolumn{8}{c}{\textbf{TLive-Omni-9B}} \\
\midrule
Method
& Opt.$\uparrow$ & State$\uparrow$
& Src.$\uparrow$ & Conf.$\uparrow$
& HR$\downarrow$ & CAR$\uparrow$ & OR$\downarrow$ \\
\midrule
SFT
& 75.71 & 62.51 & 60.26 & 88.86
& 13.69 & 86.38 & \textbf{25.97} \\
DPO
& 76.12 & 62.85 & 59.94 & 93.01 & 12.69 & 92.51 & 27.65 \\
GRPO
& 75.27 & 62.32 & 58.87 & 91.14 & 12.95 & 89.67 & 28.04 \\
mDPO
& 74.51 & 61.63 & 60.83 & 91.57 & 13.25 & 90.42 & 29.20 \\
BPO
& 74.36 & 62.23 & 61.47 & 94.48 & 11.68 & 92.92 & 30.07 \\
\rowcolor{MethodBlue}
\textbf{EI-GRPO}
& \textbf{78.12} & \textbf{69.10}
& \textbf{63.31} & \textbf{95.98}
& \textbf{11.21} & \textbf{94.61} & \textbf{25.97} \\
\bottomrule
\end{tabular*}
\end{minipage}

\end{table*}

\subsection{Main Results}
\label{sec:main_results}
\label{sec:base_model_evaluation}
\label{sec:training_comparison}

\noindent\textbf{Benchmark performance.}
Table~\ref{tab:base_response_evidence} reveals
distinct strengths in answer selection, evidence
assessment, and abstention.
GPT-5.5 leads in option accuracy at 86.83\%
but achieves only 38.50\% state accuracy,
showing that selecting correct answers does not
ensure accurate recognition of their evidence states.
Gemini 3.1 Pro, instead, leads in state accuracy
and source attribution, demonstrating that the
strongest answer selector is not necessarily
the strongest evidence assessor.
Among open-source models, Qwen3-Omni-30B performs
well in citation matching but remains weak at
conflict detection.
Conversely, TLive-Omni-9B detects conflicts better
, despite lower option accuracy and source F1.

Abstention metrics reveal another trade-off:
TLive-Omni-9B has the highest open-source CAR
and the lowest HR, but also the highest OR.

GPT-5.5 better balances these decisions, whereas
Gemini 3 Flash achieves the lowest HR but
has more unnecessary refusals.
Moreover, high CAR does not guarantee low HR,
which also captures unavailable-source citations
and incorrect factual values.Schema-valid coverage for all evaluated models and
post-training variants, together with the output
validation criteria, is reported in the supplementary
material.

\begin{figure}[!htbp]    \centering    \includegraphics[width=\linewidth]{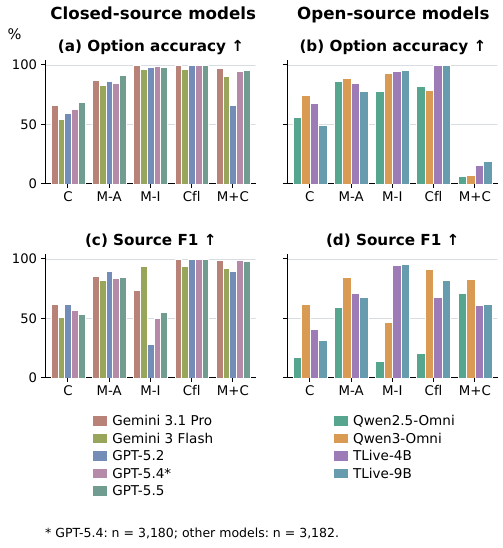}    \caption{Option accuracy and source F1 across five evidence conditions: C, complete-and-consistent; M-A, missing-but-answerable; M-I, missing-and-insufficient; Cfl, conflict-only; M+C, missing-and-conflicting. GPT-5.4 uses 3,180 scored examples; other models use 3,182. Claude is omitted because same-version condition-level results are unavailable.}    \label{fig:evidence_condition_results}\end{figure}

\noindent\textbf{Performance across evidence conditions.}Fig.~\ref{fig:evidence_condition_results} compares option accuracy and source F1 across five evidence conditions for nine models. On M+C, open-source models achieve only 6.22--19.19\% option accuracy despite source F1 of 61.02--83.14\%, showing that citation matching does not ensure reliable answer selection. The five closed-source models also perform worse on M+C than on conflict-only inputs in both metrics, highlighting the challenge of combined missingness and conflict.

\noindent\textbf{Post-training comparisons.}
Table~\ref{tab:training_comparison} shows more
consistent improvements from EI-GRPO than from
the other evaluated post-training methods.
Its gains in option accuracy are accompanied
by state accuracy improvements of 6.59-9.75
percentage points over SFT, extending the benefits
beyond response selection to evidence assessment.
Source attribution and conflict detection also
improve jointly, with the largest gains on
Qwen2.5-Omni-7B.
By contrast, DPO improves attribution on this
backbone while weakening conflict detection,
showing that better citation matching alone
does not ensure more reliable evidence interpretation.
A related trade-off appears in BPO on TLive-Omni-9B:
improved required abstention and lower HR are
accompanied by greater over-refusal.
EI-GRPO instead increases CAR and reduces HR
without increasing OR on any backbone.
Its reliability gains therefore do not come
at the cost of more refusals on answerable questions,
supporting improved discrimination between when
to answer and when to abstain.
Nevertheless, evidence-state recognition remains
less accurate than response selection, and
TLive-Omni-4B retains substantial over-refusal
despite its high CAR.
Post-training thus improves several complementary
capabilities without fully resolving evidence-aware
decision-making.

\subsection{Ablation Study}
\label{sec:ei_grpo_ablations}

\par\noindent\textbf{Settings.} We evaluate four variants. \textit{w/o field bonus} sets $\gamma=0$. \textit{w/o all-pairs} uses only same-index pairs while retaining endpoint-specific advantages. \textit{w/o endpoint credit} retains all cross-endpoint pairs but assigns a shared advantage to both endpoints of each pair. \textit{w/o field credit} replaces the field bonus with a binary bonus requiring both complete responses to be correct. Single-response rewards are retained throughout. All variants share the SFT initialization, random seed, and generation cap.\begin{table}[!t]
\centering
\caption{Ablation Study (\%). Best results per backbone
are bold.}
\label{tab:ei_grpo_ablations}
\footnotesize
\setlength{\tabcolsep}{1.2pt}
\renewcommand{\arraystretch}{1.0}
\begin{tabular*}{\columnwidth}{@{\extracolsep{\fill}}lccccccc@{}}
\toprule
\textbf{Method} & Opt.$\uparrow$ & State$\uparrow$ &
Src.$\uparrow$ & Conf.$\uparrow$ & HR$\downarrow$ &
CAR$\uparrow$ & OR$\downarrow$ \\
\midrule
\rowcolor{AblModelYellow}
\multicolumn{8}{c}{\rule[-3pt]{0pt}{10pt}\textbf{Qwen2.5-Omni-7B}} \\
\midrule
\textbf{EI-GRPO} &
\cellcolor{AblBestBlue}\textbf{75.33} &
\cellcolor{AblBestBlue}\textbf{64.66} &
\cellcolor{AblBestBlue}\textbf{69.93} &
\cellcolor{AblBestBlue}\textbf{58.87} &
\cellcolor{AblBestBlue}\textbf{18.73} &
\cellcolor{AblBestBlue}\textbf{59.13} &
\cellcolor{AblBestBlue}\textbf{15.63} \\
\textit{w/o field bonus} &
71.78 & 55.63 & 50.65 & 38.81 & 20.88 & 49.85 & 22.20 \\
	\textit{w/o all-pairs} & 72.12 & 56.16 & 49.96 & 43.96 & 20.38 & 52.10 & 22.20 \\ \textit{w/o endpoint credit} & 71.94 & 56.00 & 49.98 & 43.60 & 20.38 & 52.51 & 21.82 \\
\textit{w/o field credit} &
72.22 & 56.44 & 49.75 & 46.10 & 20.07 & 53.89 & 22.51 \\
\midrule
\rowcolor{AblModelYellow}
\multicolumn{8}{c}{\rule[-3pt]{0pt}{10pt}\textbf{Qwen3-Omni-30B}} \\
\midrule
\textbf{EI-GRPO} &
\cellcolor{AblBestBlue}\textbf{87.14} &
\cellcolor{AblBestBlue}\textbf{78.67} &
\cellcolor{AblBestBlue}\textbf{67.54} &
\cellcolor{AblBestBlue}\textbf{88.64} &
\cellcolor{AblBestBlue}\textbf{12.39} &
\cellcolor{AblBestBlue}\textbf{89.13} &
\cellcolor{AblBestBlue}\textbf{11.20} \\
\textit{w/o field bonus} &
83.11 & 67.84 & 61.56 & 85.08 & 14.09 & 84.18 & 14.20 \\
	\textit{w/o all-pairs} & 84.38 & 69.58 & 63.52 & 85.46 & 13.77 & 83.98 & 14.24 \\ \textit{w/o endpoint credit} & 83.94 & 70.71 & 62.68 & 83.57 & 14.11 & 82.78 & 14.72 \\
\textit{w/o field credit} &
82.67 & 70.77 & 62.74 & 84.51 & 13.88 & 83.08 & 13.88 \\
\midrule
\rowcolor{AblModelYellow}
\multicolumn{8}{c}{\rule[-3pt]{0pt}{10pt}\textbf{TLive-Omni-4B}} \\
\midrule
\textbf{EI-GRPO} &
\cellcolor{AblBestBlue}\textbf{78.27} &
\cellcolor{AblBestBlue}\textbf{76.12} &
\cellcolor{AblBestBlue}\textbf{71.37} &
\cellcolor{AblBestBlue}\textbf{93.04} &
\cellcolor{AblBestBlue}\textbf{11.76} &
\cellcolor{AblBestBlue}\textbf{98.80} &
\cellcolor{AblBestBlue}\textbf{27.28} \\
\textit{w/o field bonus} &
75.52 & 66.25 & 68.54 & 90.71 & 12.67 & 91.47 & 29.24 \\
	\textit{w/o all-pairs} & 75.64 & 66.59 & 68.62 & 91.37 & 12.57 & 92.22 & 29.20 \\ \textit{w/o endpoint credit} & 75.39 & 66.40 & 68.56 & 90.33 & 12.77 & 91.17 & 29.12 \\
\textit{w/o field credit} &
75.39 & 66.47 & 68.58 & 90.17 & 12.86 & 90.87 & 29.12 \\
\midrule
\rowcolor{AblModelYellow}
\multicolumn{8}{c}{\rule[-3pt]{0pt}{10pt}\textbf{TLive-Omni-9B}} \\
\midrule
\textbf{EI-GRPO} &
\cellcolor{AblBestBlue}\textbf{78.12} &
\cellcolor{AblBestBlue}\textbf{69.10} &
\cellcolor{AblBestBlue}\textbf{63.31} &
\cellcolor{AblBestBlue}\textbf{95.98} &
\cellcolor{AblBestBlue}\textbf{11.21} &
\cellcolor{AblBestBlue}\textbf{94.61} &
25.97 \\
\textit{w/o field bonus} &
76.21 & 63.01 & 60.63 & 90.42 & 13.17 & 88.32 &
\cellcolor{AblBestBlue}\textbf{25.66} \\
	\textit{w/o all-pairs} & 76.15 & 63.01 & 60.58 & 90.70 & 13.26 & 88.62 & 25.82 \\ \textit{w/o endpoint credit} & 75.61 & 62.57 & 60.18 & 89.58 & 13.54 & 87.57 & 26.25 \\
\textit{w/o field credit} &
75.93 & 62.76 & 60.48 & 90.00 & 13.41 & 87.87 & 25.97 \\
\bottomrule
\end{tabular*}
\end{table}

\par\noindent\textbf{Results.}\quad
Table~\ref{tab:ei_grpo_ablations} separates the effects of pairing and credit assignment. Relative to EI-GRPO, same-index pairing reduces option accuracy by 1.97--3.21 points, while shared advantages reduce it by 2.51--3.39 points. Both variants also reduce state accuracy, source F1, conflict F1, and CAR and increase HR on all four backbones, supporting all-pairs aggregation and endpoint-specific credit assignment. OR is not uniformly improved: same-index pairing lowers it by 0.15 points on TLive-9B. Removing the field bonus reduces source and conflict F1 by 19.28 and 20.06 points on Qwen2.5; replacing it with a binary bonus lowers state accuracy across all backbones, supporting fine-grained rewards.

\section{Conclusion}

EviBench evaluates answering, evidence attribution, and abstention under missing and conflicting evidence. By distinguishing source absence from evidence insufficiency, the benchmark tests whether available evidence supports answering or requires abstention. Across ten MLLMs, correct answers do not ensure reliable evidence assessment. EI-GRPO combines cross-condition field-correctness rewards with endpoint-specific credit assignment to learn from paired evidence inputs requiring potentially different outputs. Across four backbones, it improves answer selection, source attribution, and conflict detection over SFT while reducing structured hallucination without increasing aggregate over-refusal. Separate ablations support the contributions of all-pairs aggregation and endpoint-specific credit assignment within this training framework. Evidence-state recognition and over-refusal remain challenges. Our findings are limited to controlled interventions and structured evaluation; robustness in unconstrained livestream conversations requires further validation. Future work will study clarification and evidence acquisition in evolving livestreams.

\section*{LLM Usage Statement}
\looseness=-1
GPT-5.5 and Gemini 3.1 Pro were used to generate questions,
four candidate responses per question, and reference decisions
for EviBench, as described in Sec.~\ref{sec:data_construction}.
All samples generated by these models were manually verified and corrected where necessary before inclusion. The multimodal
language models listed in Sec.~\ref{sec:experimental_setup}
were used as evaluation subjects and, for the four open-source
backbones, in the post-training experiments described therein.
GPT-6 was used to polish the language and expression of the
manuscript.

\clearpage
\setcounter{table}{0}
\setcounter{figure}{0}
\setcounter{equation}{0}
\renewcommand{\thetable}{S\arabic{table}}
\renewcommand{\thefigure}{S\arabic{figure}}
\renewcommand{\theHequation}{supp.\arabic{equation}}
\renewcommand{\theHtable}{supp.\arabic{table}}
\renewcommand{\theHfigure}{supp.\arabic{figure}}

\raggedbottom
\makeatletter
\setlength{\@fptop}{0pt}
\setlength{\@dblfptop}{0pt}
\setlength{\@dblfpsep}{12pt}
\setlength{\@dblfpbot}{0pt plus 1fil}
\makeatother
\renewcommand{\arraystretch}{1.12}
\lstset{basicstyle=\rmfamily\small,breaklines=true,breakatwhitespace=false,breakautoindent=false,breakindent=1em,columns=fullflexible,keepspaces=true,showstringspaces=false,frame=l,framerule=0.5pt,rulecolor=\color{black!25},backgroundcolor=\color{black!2},framesep=7pt,xleftmargin=10pt,xrightmargin=7pt,aboveskip=10pt,belowskip=10pt}

\section*{APPENDIX}
\appendix

\noindent This appendix provides additional details on data construction and partitioning, human verification, evaluation protocols, and the implementation of EI-GRPO. It also includes supplementary diagnostics, hyperparameter experiments, and the evaluation prompt.

\section{Data Sources and Dataset Splits}
\label{sec:data}
\subsection{Source records and counting units}
EviBench is organized around records linking product information to livestream material. A source record, indexed by \field{query\_id}, can contain a product title, attributes, product images, video or sampled frames, audio, and an ASR transcript. The questions are in Chinese. A source record is not necessarily a distinct product or an independent livestream session: the same product can occur in multiple clips. Likewise, a recorded \field{video\_id} should not be interpreted as a verified full livestream-session identifier.

Table~\ref{tab:source} reports the observed identifier counts together with their coverage. Coverage is the number of endpoint inputs in which the field is present; the distinct count is computed only over those inputs.

\begin{table*}[t]
\centering\small
\caption{Source identifiers in the supplied data inventory. Each identifier entry is ``distinct identifiers / covered endpoint inputs.'' The test-set counts reflect inventory availability, not necessarily the normalized identifiers used in the overlap audit.}
\label{tab:source}
\begin{tabular*}{\textwidth}{@{\extracolsep{\fill}}lrrrr@{}}
\toprule
Split & Inputs/questions & Query ID & Product ID & Video ID \\
\midrule
SFT & 7,996 & 931 / 7,996 & 563 / 5,536 & 931 / 7,996 \\
RL & 7,897 & 930 / 7,897 & 578 / 5,550 & 930 / 7,897 \\
Validation & 3,408 & 393 / 3,408 & 254 / 2,394 & 393 / 3,408 \\
EviBench & 3,182 & 1,160 / 2,925 & 629 / 1,739 & 856 / 1,737 \\
\bottomrule
\end{tabular*}
\end{table*}

The reinforcement-learning pool contains 7,897 original questions. These are distinct from the candidate responses sampled during policy optimization. The number of original questions should not be conflated with the number of sampled responses or optimization steps.

The test inventory contains 1,160 distinct query identifiers among 2,925 of its 3,182 inputs. This observation alone does not reconstruct the complete expansion from initial source material to the final benchmark. A source-to-question-to-variant ledger is needed to associate generation, filtering, and correction counts with a common version.

\subsection{Tasks and evidence conditions}
The benchmark covers product inquiry, attribute identification, selling-point understanding, recommendation rationale, and after-sales questions. These tasks concern, respectively, general product information, specific properties, evidence-supported selling points, the basis of a recommendation, and issues such as sizing or care. Table~\ref{tab:distribution} gives the complete task-by-condition distribution.

\begin{table*}[t]
\centering\small
\caption{Distribution of the 3,182 EviBench examples. Complete denotes complete-and-consistent evidence; M-A denotes missing-but-answerable evidence; M-I denotes missing-and-insufficient evidence; Conflict denotes conflict without missingness; M+C combines missingness and conflict.}
\label{tab:distribution}
\begin{tabular*}{\textwidth}{@{\extracolsep{\fill}}lrrrrrr@{}}
\toprule
Task & Complete & M-A & M-I & Conflict & M+C & Total \\
\midrule
Product inquiry & 202 & 431 & 115 & 10 & 74 & 832 \\
Attribute identification & 115 & 476 & 76 & 8 & 157 & 832 \\
Selling-point understanding & 119 & 397 & 31 & 8 & 83 & 638 \\
Recommendation rationale & 114 & 185 & 12 & 12 & 32 & 355 \\
After-sales QA & 244 & 231 & 8 & 18 & 24 & 525 \\
\midrule
Total & 794 & 1,720 & 242 & 56 & 370 & 3,182 \\
\bottomrule
\end{tabular*}
\end{table*}

The five input conditions map to four decision states. Complete inputs map to \textrm{supported}, M-A to \textrm{missing}, and M-I to \textrm{insufficient}. Both Conflict and M+C map to \textrm{conflict} when the contradiction is question-relevant and unresolved. Thus, there are 2,514 answerable examples and 668 examples requiring abstention, including 426 with relevant conflict. Missingness describes source availability; insufficiency describes whether the supplied evidence supports a decision. They are not interchangeable.

\subsection{Source grouping and partitioning}
The training construction records start from a pool of 3,869 sources. Query, product, and parent-video associations are connected transitively, and the resulting groups are partitioned before expansion. The recorded allocation is 931 sources for SFT, 931 for RL, and 393 for validation; another 1,614 are isolated because of associations with a historical test collection. These four counts sum to 3,869. The corresponding SFT, RL, and validation association-group counts are 471, 478, and 140.

\section{Human Verification Protocol}
\label{sec:review}
\subsection{Generation and review workflow}
\begin{figure*}[!t]
\centering
\includegraphics[width=\textwidth]{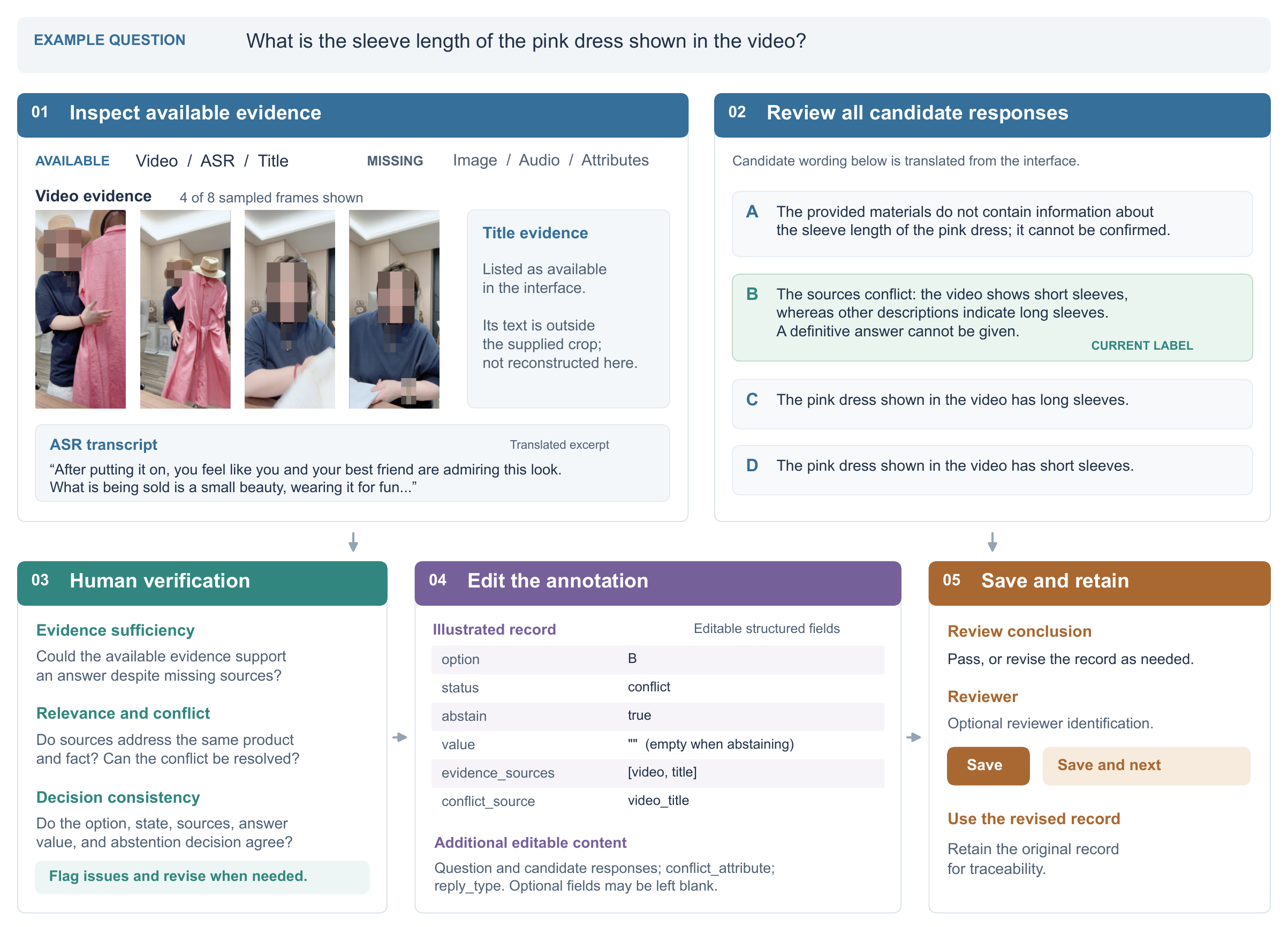}
\caption{Human verification workflow reconstructed from the annotation interface. Modules 1--2 present the available evidence and four candidate responses; modules 3--5 cover evidence checking, annotation editing, and saving with the original record retained. Faces and identifiable brand regions are pixelated. The example reproduces the displayed annotation: option B, state \field{conflict}, \field{abstain=true}, an empty \field{value}, and the conflict-source pair \field{video\_title}. The title is listed as available but its text is outside the supplied screenshot; the figure illustrates the review workflow rather than independently establishing the example's conflict label.}
\label{fig:review-workflow}
\end{figure*}

GPT and Gemini are used to propose questions, four candidate responses, and structured reference decisions. Figure~\ref{fig:review-workflow} summarizes the human-review workflow in five modules. Reviewers first inspect the question and the evidence actually provided, including available and missing sources, sampled video frames, and textual evidence (module 1), and examine all four candidate responses (module 2). They then assess evidence sufficiency, relevant conflicts, and consistency among the decision fields (module 3).

Reviewers can revise the question, candidate responses, selected option, evidence state, abstention flag, factual value, cited sources, conflicting source pair, and auxiliary annotation fields (module 4). They record the review conclusion and save the revised annotation, optionally adding a reviewer identifier; the original record is retained for traceability (module 5). The diagram reorganizes the Chinese interface into an English illustration, using four sampled frames and a translated ASR excerpt. It illustrates the procedure rather than quantifying annotation accuracy or inter-reviewer agreement.

\subsection{Evidence sufficiency, relevance, and conflict adjudication}
Review is conditioned on the sources actually delivered for the current example. A missing source does not require abstention if other available evidence sufficiently answers the question. Conversely, merely making a source available does not justify citing it: the source must support the queried fact. Candidate responses are possible decisions, not independent evidence for their own factual claims.

The minimum sufficient evidence set is distinguished from the set of acceptable citations. A minimum set establishes answerability, whereas additional verified and relevant support may also be cited. Arbitrary supersets are not automatically valid. The scoring and reward functions compare predictions with annotated acceptable sets; they do not perform an independent semantic assessment of every newly proposed citation.

In addition to evidence sufficiency and relevance, reviewers assess whether the sources conflict. A conflict concerns mutually incompatible evidence about the same queried fact, product, and specification, with no available basis for resolving the discrepancy. Different SKU variants, different times, or multiple nonexclusive attribute labels do not by themselves establish an unresolved conflict. Reviewers check product identity and the scope of each claim before assigning a conflict label.

For a relevant unresolved conflict, the reference specifies the conflicting source pair, includes those sources in the citation set, sets \field{abstain=true}, and leaves \field{value} empty. For insufficient evidence without an identified relevant contradiction, the state is \field{insufficient} and the conflict source is \field{none}. Unrelated discrepancies do not override otherwise sufficient evidence.

\section{Implementation and Evaluation Details}
\label{sec:implementation}
\subsection{Input construction and inference}
The open-source evaluation entry point fixes the 3,182-example benchmark file and its SHA-256 identifier. Two complete model replicas process disjoint shards, which are merged using unique sample IDs. This procedure is parallel inference over one evaluation set, not two stochastic runs of each example.

The recorded evaluation budget allows at most four product images and eight video frames per example, with a maximum of 256 generated tokens. Audio-capable models receive the full supplied audio. Audio and ASR remain separate sources: supplying ASR text must not be described as supplying native audio. Per-model processor transformations, frame timestamps, image resolution, and native-audio support belong to the corresponding run manifests. These details cannot be inferred uniformly from an older default configuration.

\subsection{Output schema and decision consistency}
All model responses follow a seven-field JSON protocol: \field{option}, \field{status}, \field{value}, \field{evidence\_sources}, \field{conflict\_source}, \field{abstain}, and \field{confidence}. The reference annotations are excluded from the model input.

\begin{table*}[t]
\centering\small
\caption{Field-level output protocol. Cross-field consistency and factual correctness are distinct from field-level validity.}
\label{tab:schema}
\begin{tabularx}{\textwidth}{l l X}
\toprule
Field & JSON type & Permitted content \\
\midrule
\textrm{option} & String & Exactly A, B, C, or D; it cannot be the answer text or an empty string. \\
\textrm{status} & String & supported, missing, insufficient, or conflict. The missing state remains answerable. \\
\textrm{value} & String & A short supported fact or attribute value. Empty for abstention; it may also be empty for a non-attribute response without a separately annotated value. \\
\textrm{evidence\_sources} & String array & Unique source names from image, title, parameter, asr, audio, and video. An empty array is syntactically permitted; correctness is reference-dependent. \\
\textrm{conflict\_source} & String & none, or two distinct source names joined by an underscore. Pair order is normalized for comparison. \\
\textrm{abstain} & Boolean & JSON true or false, rather than a quoted string. \\
\textrm{confidence} & Number & A finite value in $[0,1]$ expressing confidence in the complete decision, including a correct abstention. \\
\bottomrule
\end{tabularx}
\end{table*}

For \field{supported} and \field{missing}, abstention is false and the conflict source is none. For \field{insufficient}, abstention is true, the value is empty, and the conflict source is none. For \field{conflict}, abstention is true, the value is empty, and a specific conflicting pair is required; both sources should occur in the citation set. These are decision-consistency requirements, not a license to remove inconsistent predictions from every metric.

\clearpage \twocolumn \subsection{Structured hallucination and coverage}
Let $\mathcal V$ be the set of schema-valid predictions under the output validator. For each $i\in\mathcal V$, let $u_i$ indicate answering, or returning a nonempty factual value, when the reference requires abstention. Let $s_i$ indicate citation of a source unavailable in the evaluated input, and let $w_i$ indicate a nonempty incorrect value where a reference value is scored. We report \begin{equation} \begin{aligned} \mathrm{HR} &= \frac{1}{|\mathcal V|}\sum_{i\in\mathcal V} \mathbf{1}[u_i \lor s_i \lor w_i],\\[6pt] \mathrm{Schema} &= \frac{|\mathcal V|}{N}. \end{aligned} \end{equation}
Overlapping violations count once. Schema-invalid outputs are excluded from the HR denominator, not counted as hallucination-free. HR captures these structured violations rather than an exhaustive semantic judgment of every candidate response or any additional text. An available but irrelevant citation may escape this metric, and a wrong option is not automatically a hallucination. HR should therefore be interpreted together with Schema, answering accuracy, and OR.

Table~\ref{tab:coverage} reports Schema and HR for Base, SFT, and EI-GRPO on each of the four backbones. Each entry uses $N=3,182$ predictions, with HR computed over schema-valid outputs. The EI-GRPO HR values match the main paper; count-based diagnostics for these runs will be completed once the corresponding records are confirmed.

\begin{table*}[t]
\centering\small
\caption{Schema-valid output counts and rates, and structured hallucination rates, for Base, SFT, and EI-GRPO. All entries use $N=3,182$ predictions. Base denotes the original backbone. EI-GRPO HR values follow the main-paper results. HR is averaged over multiple evaluation runs and is therefore not necessarily equal to an integer violation count divided by the displayed valid-output count.}
\label{tab:coverage}
\begin{tabular*}{\textwidth}{@{\extracolsep{\fill}}llrrr@{}}
\toprule
Model & Method & Valid outputs & Schema (\%) & HR (\%) \\
\midrule
Qwen2.5-Omni & Base & 3167 & 99.53 & 32.33 \\
 & SFT & 3172 & 99.69 & 20.11 \\
 & EI-GRPO & 3168 & 99.56 & 18.73 \\
\midrule
Qwen3-Omni & Base & 3178 & 99.87 & 28.01 \\
 & SFT & 3165 & 99.47 & 14.06 \\
 & EI-GRPO & 3168 & 99.56 & 12.39 \\
\midrule
TLive-4B & Base & 3133 & 98.46 & 23.33 \\
 & SFT & 3166 & 99.50 & 12.63 \\
 & EI-GRPO & 3166  & 99.50 & 11.76 \\
\midrule
TLive-9B & Base & 3081 & 96.83 & 21.49 \\
 & SFT & 3170 & 99.62 & 13.69 \\
 & EI-GRPO & 3171 & 99.65 & 11.21 \\
\bottomrule
\end{tabular*}
\end{table*}

\subsection{Supervised initialization and final configuration}
SFT supervises assistant response tokens while masking prompt tokens, media placeholders, and the numerical confidence placeholder. Multimodal encoders remain frozen. The documented SFT configuration uses 7,996 endpoint inputs, learning rate $10^{-5}$, LoRA rank 16, scaling 32, dropout 0.05, microbatch size one, gradient accumulation four, and seed 42. The recorded entry-point defaults include maximum sequence length 16,384, a pixel budget of 200,704, BF16 precision, cosine scheduling, warm-up ratio 0.03, and gradient clipping at 1.0. The actual language-side target modules are selected per backbone and recorded separately.

For the final configuration specified for the main paper, EI-GRPO uses two candidates per endpoint, source-field weight $w_{\mathrm{src}}=0.40$, reference coefficient $\beta=0.10$, cross-condition coefficient $\gamma=0.25$, and policy clipping $\epsilon=0.20$. The five remaining field weights share the residual weight equally, giving 0.12 each. These field weights differ from the base-reward component weights.

The archived EI-GRPO implementation uses language-side LoRA, learning rate $10^{-6}$, AdamW with zero weight decay, dropout zero, seed 42, gradient clipping 1.0, and two policy-update passes per rollout. Sampling uses temperature one, top-$p$ one, top-$k$ zero, and at most 256 generated tokens. The SFT checkpoint initializes both the trainable policy and the frozen reference. The old sampling policy provides the importance-sampling probabilities, while the reference remains fixed.

\section{Additional Validation Experiments}
\label{sec:sensitivity}
\begin{figure*}[!t]
\centering
\includegraphics[width=\textwidth]{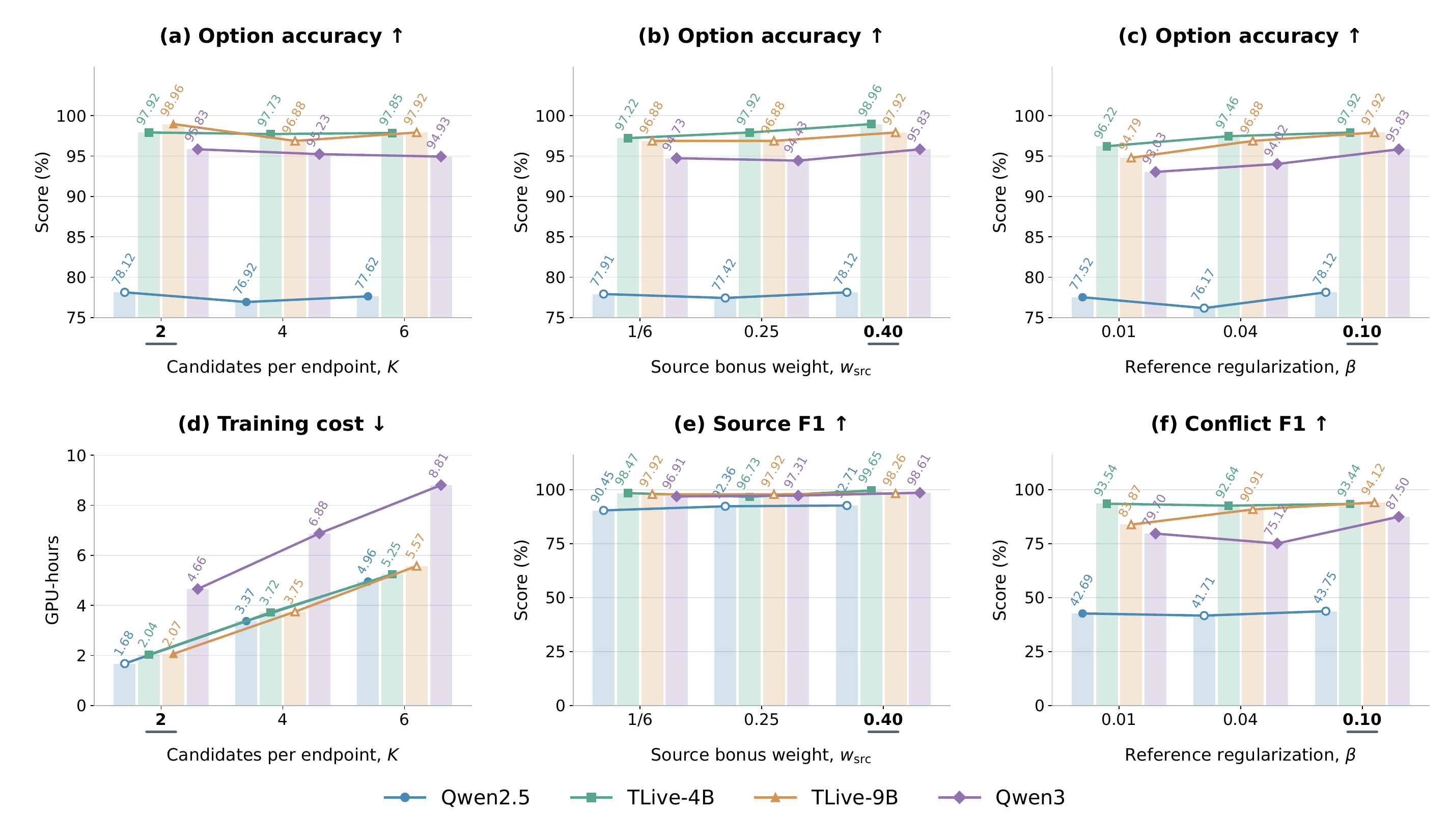}
\caption{Hyperparameter experiments. Columns show the candidate-count sweep (a, d), source-field-weight sweep (b, e), and reference-coefficient sweep (c, f). The top row reports option accuracy; the bottom row reports training cost, source F1, and conflict F1, respectively. Each sweep varies one parameter; performance scores are averaged over multiple runs. The Qwen3 training cost at four candidates per endpoint is 6.88 GPU-hours.}
\label{fig:sensitivity}
\end{figure*}

\subsection{Screening protocol}
Figure~\ref{fig:sensitivity} presents the hyperparameter results confirmed for this study. These validation results are reported separately from the 3,182-example EviBench test-set evaluation. The plotted performance scores are averaged over multiple runs, rather than reported as single-run accuracies. Fixed data exposure does not imply equal computation.

The baseline for these one-factor sweeps is $K=4$, $w_{\mathrm{src}}=0.25$, $\beta=0.04$, and $\gamma=0.25$. We inspect $K\in\{2,4,6\}$, $w_{\mathrm{src}}\in\{1/6,0.25,0.40\}$, and $\beta\in\{0.01,0.04,0.10\}$. When the source-field weight changes, the other five fields share the remaining mass equally. Archived experiments also vary $\gamma$ and base-reward weights.

\subsection{Candidate count and computation}
Increasing the candidate count does not consistently improve option accuracy. For $K=2,4,6$ candidates per endpoint, Qwen2.5 obtains 78.12\%, 76.92\%, and 77.62\%, while its training cost rises from 1.68 to 3.37 and 4.96 GPU-hours. TLive-4B achieves 97.92\%, 97.73\%, and 97.85\%, with costs of 2.04, 3.72, and 5.25 GPU-hours. TLive-9B achieves 98.96\%, 96.88\%, and 97.92\%, with costs of 2.07, 3.75, and 5.57 GPU-hours. Qwen3 achieves 95.83\%, 95.23\%, and 94.93\%, with costs of 4.66, 6.88, and 8.81 GPU-hours. In this sweep, two candidates per endpoint provide the highest option accuracy and lowest measured training cost for each backbone.

\clearpage

\end{document}